\documentclass[10pt,twocolumn,letterpaper]{article}

\usepackage[pagenumbers]{cvpr} 

\usepackage{algorithm2e}
\SetKwComment{Comment}{/* }{ */}
\RestyleAlgo{ruled}
\usepackage{multirow}
\usepackage{multicol}
\usepackage{caption}
\usepackage{subcaption}
\usepackage{dsfont}
\usepackage{courier}
\usepackage{listings}
\definecolor{cvprblue}{rgb}{0.21,0.49,0.74}
\usepackage[pagebackref,breaklinks,colorlinks,allcolors=cvprblue]{hyperref}

\def\paperID{599} 
\def\confName{CVPR}
\def\confYear{2025}

\title{Provoking Multi-modal Few-Shot LVLM via Exploration-Exploitation  \\  In-Context Learning}

\author{Cheng Chen$^{*~1,2}$, Yunpeng Zhai$^{*~2}$, Yifan Zhao$^{\dag~1}$, Jinyang Gao$^2$, Bolin Ding$^2$, Jia Li$^{\dag~1}$\\
$^1$State Key Laboratory of Virtual Reality Technology and Systems \\
$^2$Tongyi Lab, Alibaba Group\\
{\tt\small \{chengchen, zhaoyf, jiali\}@cvteam.net} \\
{\tt\small \{zhaiyunpeng.zyp, jinyang.gjy, bolin.ding\}@alibaba-inc.com}
}

\begin{document}
\maketitle
\def\thefootnote{*}\footnotetext{These authors contributed equally to this work.}\def\thefootnote{\arabic{footnote}}
\def\thefootnote{$\dag$}\footnotetext{Corresponding authors.}\def\thefootnote{\arabic{footnote}}
\begin{abstract}
    In-context learning (ICL), a predominant trend in instruction learning, aims at enhancing the performance of large language models by providing clear task guidance and examples, improving their capability in task understanding and execution. This paper investigates ICL on Large Vision-Language Models (LVLMs) and explores the policies of multi-modal demonstration selection. Existing research efforts in ICL face significant challenges: First, they rely on pre-defined demonstrations or heuristic selecting strategies based on human intuition, which are usually inadequate for covering diverse task requirements, leading to sub-optimal solutions; Second, individually selecting each demonstration fails in modeling the interactions between them, resulting in information redundancy. Unlike these prevailing efforts, we propose a new exploration-exploitation reinforcement learning framework, which explores policies to fuse multi-modal information and adaptively select adequate demonstrations as an integrated whole. The framework allows LVLMs to optimize themselves by continually refining their demonstrations through self-exploration, enabling the ability to autonomously identify and generate the most effective selection policies for in-context learning.
    Experimental results verify the superior performance of our approach on four Visual Question-Answering (VQA) datasets, demonstrating its effectiveness in enhancing the generalization capability of few-shot LVLMs.
\end{abstract}

\section{Introduction}
\label{sec:intro}

\begin{figure}[t]
  \centering
   \includegraphics[width=\linewidth]{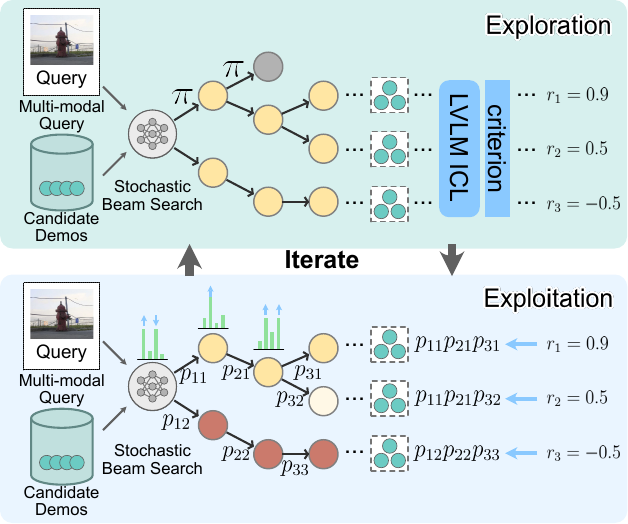}
   \caption{The motivation of the proposed method. The method consists of two stages: exploration and exploitation. In the exploration stage, a stochastic beam search is conducted to generate multiple selection strategies for the query. We reason the output with a large language model and comprehensively evaluate these multiple demonstration sets. In the exploitation stage, the policy is optimized based on evaluation, making it more likely to select the optimal combination. These two stages iterate continuously, resulting in seeking the optimal strategy.}
   \label{fig:motivation}
\end{figure}

Recent years have witnessed remarkable advancements in Large Vision-Language Models (LVLMs)~\cite{zhao2024surveylargelanguagemodels}. Powered by the rapid development of Language Models (LMs) and extensive pretraining on large-scale vision-language datasets, LVLMs have demonstrated exceptional capabilities on various tasks, including generation~\cite{du2022glm, touvron2023llamaopenefficientfoundation, Bai2023qwen, anil2023palm2technicalreport, openai2024gpt4technicalreport}, translation \cite{he2024exploring, li2024eliciting, Yin2023Gloss, Li2022VALHALLA, Gupta2023CLIPTrans}, and beyond \cite{wushi2022adversarial, guo2024multimodal, gou2023mvp, ling2022vision}. Despite their impressive performance, LVLMs still encounter substantial challenges in adapting to novel scenarios, primarily due to the resource-intensive nature of fine-tuning procedures.

To tackle this problem, In-Context Learning (ICL) offers a training-free alternative to conventional fine-tuning approaches. Specifically, ICL augments the prompt with a few high-quality examples, \ie, demonstrations, allowing models to learn from them efficiently during inference. This enables LVLMs to leverage analogical reasoning and effectively generalize to novel tasks. Consequently, ICL is becoming a widely-used technique for easily improving the performance of both LLMs and VLMs \cite{wang2022training, zemlyanskiy2022generate, Itay2023Diverse, gao2024ambiguityawareincontextlearninglarge, nandigam2022diverse}.

The selection of demonstrations plays a crucial role in ICL for LVLMs. Current approaches predominantly rely on heuristic selection strategies, which can be categorized into two main lines. The first line adopts query-independent selection, utilizing predefined demonstrations for simplicity and stability. However, this static approach fails to cover the diverse range of task requirements, leading to suboptimal performance when handling novel vision-language tasks \cite{Brown2020Language,wang2022training,zemlyanskiy2022generate}. The second line employs dynamic selection strategies, typically retrieving demonstrations with higher embedding similarity to the query. While similarity-based methods have shown promise in language-only domains, they face significant limitations in multi-modal scenarios. 
First, high similarity scores {do not guarantee high demonstration utility}, potentially overlooking valuable demonstrations while retrieving samples with limited reference value. Second, the individual selection of demonstrations results in {substantial information redundancy among retrieved top-similar samples}, since it doesn't consider the inter-sample relationships. This severely hampers the model's generation capability. 
\textbf{These heuristic approaches, whether through manual curation or fixed strategies, suffer from a lack of learning capability, failing to fully exploit appropriate demonstrations and thus severely limiting the potential of ICL.}
Moreover, the inherent difficulty of vision-language alignment poses additional challenges for demonstration selection, as it requires effectively capturing both visual and textual semantic relationships.

This paper investigates the multi-modal demonstration selection for in-context learning of LVLMs. 
Departing from heuristic selection approaches, we introduce an end-to-end learnable framework that automatically optimizes the demonstration combinations.
To address the redundancy issues, we reformulate demonstration selection as a combinatorial optimization problem rather than independent sample selection. However, unlike conventional supervised learning tasks, this combinatorial optimization problem is particularly challenging as the optimal demonstration combinations cannot be manually annotated, resulting in the absence of direct supervision signals. To address the challenge, we propose a novel exploration-exploitation framework that adaptively learns to choose demonstrations in a reinforcement-like manner.
As illustrated in \Cref{fig:motivation}, our framework iteratively conducts an exploration phase to efficiently explore good demonstration combinations and an exploitation phase to reinforce the sampling-policy network. Given a query (image-question pair) and a candidate demonstration set, the sampling-policy network first leverages interaction encoding to establish cross-modal relationships, then performs auto-regressive demonstration selection, where each selection decision is informed by the ensemble of previously selected samples.
In the exploration phase, the framework applies stochastic beam search to the sampling-policy network, generating multiple possible demonstration combinations. These combinations are then evaluated by feeding them along with the query into the LVLM to obtain effective rewards. In the exploitation phase, the sampling policy is optimized using policy gradient reinforcement learning based on the rewards of different combinations. This process strengthens the joint probability of high-performing combinations while suppressing that of less effective ones.
Through alternating between exploration and exploitation phases, the framework enables models to autonomously identify and generate the most effective demonstration sets for input queries. This approach reduces the dependence on manual intervention and significantly improves generalization and adaptability.

The contributions of this paper are summarized below:
\begin{itemize}
  \item We view ICL demonstration selection from a new perspective of combinatorial optimization. It mitigates the drawbacks of individual selection by interaction among samples, thereby reducing information redundancy and enhancing demo utility.
  \item We introduce a multi-modal exploration-exploitation framework. It iteratively conducts an exploration phase to explore better demonstration combinations by stochastic tree search, and an exploitation phase to adaptively optimize the sampling-policy network in a reinforcement manner.
  \item Extensive experiments on four representative VQA datasets across multiple LVLMs demonstrate the effectiveness of our methods.
\end{itemize}
\begin{figure*}[t]
    \centering
     \includegraphics[width=\linewidth]{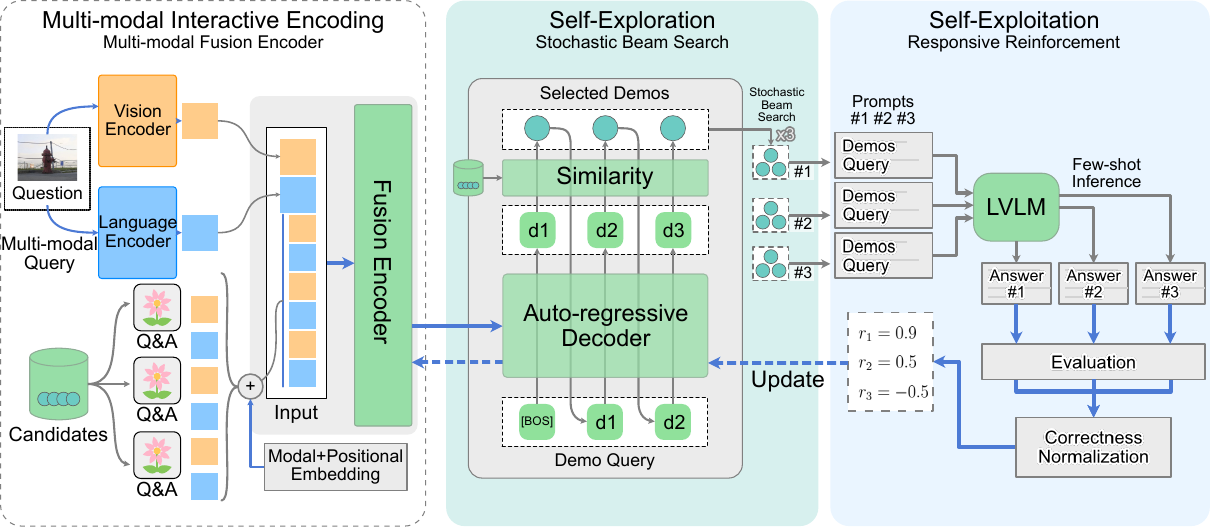}
     \caption{The pipeline of the proposed framework. It comprises three parts: multi-modal interactive encoding, self-exploration, and self-exploitation. First, the questions and images of queries and candidate demonstrations are encoded to generate multi-modal features, which are subsequently processed through a fusion encoder to output the interactive features. Following this, we employ auto-regressive decoder to generate multiple demonstration combinations that are suitable for the query. These combinations are used in few-shot inference. Finally, the predicted outputs are compared against the ground-truth to get responsive rewards, leveraged to optimize the selection policy.}
     \label{fig:onecol}
  \end{figure*}
\section{Related Works}

\subsection{In-context Learning}

Pre-trained VLMs show the ability to understand and reason. Consequently, in-context learning (ICL) uses examples related to the input query to activate the few-shot ability of models and predict the answers through analogy. Many methods have been explored. Brown \etal \cite{Brown2020Language} first explore the few-shot ability of the language models and propose in-context learning with fixed demonstrations.
Recent works have shown that dynamical retrieval leads to better improvement compared with the fixed. The dynamical retrieval shares similar ideas with retrieval-augmented generation (RAG), for example, MuRAG \cite{chen2022murag} selects multi-modal references related to questions, while the different motivations lead to different optimization goals and evaluations. Wang \etal \cite{wang2022training} propose a method that retrieves instances similar to the query. Zemlyanskiy \etal \cite{zemlyanskiy2022generate} also propose retrieval methods based on similarity but it considers both input and output.
Besides similarity, diversity has been proven effective \cite{Itay2023Diverse,nandigam2022diverse,gao2024ambiguityawareincontextlearninglarge}. For example, Itay \etal \cite{Itay2023Diverse} find that various structures encourage the model to generalize and propose methods to select diverse demonstrations. Gao \etal \cite{gao2024ambiguityawareincontextlearninglarge} find that ambiguous examples can better improve ICL performance compared to selecting semantically similar ones only. Li \etal \cite{li2023unified} propose a unified retriever by training a model on multiple tasks with ranking loss.
While most works focus on LLMs, we extend the research to the LVLMs and further investigate efficient information utilization for better demonstration combination strategies.

\subsection{Large Vision-Language Model}

Building upon the success of LLMs, researchers are focusing on large vision-language models (LVLMs) and have achieved substantial advancements, which enable the understanding of complex interactions between images and text \cite{zhang2024mm}.
GPT-4 (Vision) \cite{openai2024gpt4technicalreport} and Gemini \cite{geminiteam2024geminifamilyhighlycapable} first explore the region, showing impressive multi-modal understanding and generation capabilities.
Liu \etal \cite{Liu2023llava} train a large multimodal model LLaVA that connects vision and language for general understanding.
Similar work includes Minigpt-4 \cite {zhu2023minigpt4enhancingvisionlanguageunderstanding}, BLIP-2 \cite{Li2023blip2} and others.
Later, new work explores how to enable LVLMs to predict multimodal outputs, such as Koh \etal \cite{Koh2023generating} and Zheng \etal \cite{zheng2024minigpt5interleavedvisionandlanguagegeneration}, they enable the LLMs to predict visual outputs with the text representations.
Early ICL work primarily focuses on LLMs. With the development of LVLMs, there has been growing attention on how to utilize multimodal information and propose more effective ICL strategies on LVLMs.

\subsection{LLM with Reinforcement Learning}

In the realm of reinforcement learning, the concepts of exploration and exploitation are the foundation to make decisions.
Recent advancements in exploration and exploitation strategies have led to several notable algorithms.
Schulman \etal \cite{schulman2017proximalpolicyoptimizationalgorithms} propose an effective PPO algorithm, which allows for effective exploration while ensuring that the agent does not deviate far from a learned policy.
Rafailov \etal \cite{rafailov2024directpreferenceoptimizationlanguage} simplify the complex RL algorithm and propose DPO to solve the standard RLHF problem by using a simple classification loss.
Later, Shao \etal \cite{shao2024deepseekmathpushinglimitsmathematical} introduce a variant of PPO, \ie, GRPO, replacing value function approximation with the average reward of multiple samples.
These methods exemplify the significant progress in balancing exploration and exploitation, substantially advancing the efficiency of reinforcement learning.
\section{Methods}

\subsection{ Formulation and Overview}

Denote the tasks $\mathcal T$, $(\mathbf t,\mathcal I,\mathbf y) \sim \mathcal T$ a query with answer, and the large vision-language model as $\mathcal F:\mathbf x, \mathcal I \mapsto \mathbf y$ that receives a prompt $\mathbf t$ and its associated RGB images sequence $\mathcal I=\{\mathbf I_1,\mathbf I_2,\dots\}$, where $\mathbf I_{(\cdot)} \in \mathbb R^{CHW}$, to predict the textual output $\mathbf y$.

Denote $\mathcal D$ as the candidate demonstrations, our few-shot in-context learning tries to find an adaptive policy
\begin{equation}
\mathcal P:\mathbf t_q, \mathcal I_q \mapsto \mathcal E
\end{equation}
to select optimal demonstration \textbf{combination}
\begin{equation}
 \mathcal E=\{\mathbf d_1, \mathbf d_2,\dots,\mathbf d_m \} \subset \mathcal D
\end{equation}
for a query $(\mathbf t_q,\mathcal I_q)$, where $\mathbf d_i=(\mathbf t_i,\mathcal I_i,\mathbf y_i)$ is a demonstration and $\mathbf y$ is the associated answer to its query $\mathbf t$. With $\mathcal E$, VLM $\mathcal F$ predicts the improved output by feeding an input $P(\mathbf t_q, \mathcal E)$ of the form
\begin{equation}
 P(\mathbf t_q,\mathcal E)=\mathbf t_1 \quad \mathbf  y_1 \quad \mathbf t_2 \quad \mathbf  y_2 \quad \dots \quad \mathbf t_m \quad \mathbf  y_m \quad \mathbf t_q.
\end{equation}
To sum up, our objective is to find a policy for demonstration selection:
\begin{equation}
 \mathcal P_*={\arg\max}_{\mathcal P}\mathbb E_{(\mathbf t,\mathcal I,\mathbf y) \sim \mathcal T} \xi(\mathcal F(P(\mathbf t_q, \mathcal E),\cup \mathcal I),\mathbf y),
\end{equation}
where $\cup \mathcal I$ contain all images from demonstrations and the query, and $\xi$ is the evaluation criterion, \eg, VQAScore.

\textbf{Motivation and Framework}
As aforementioned, the interactions between demonstrations matter in ICL, and multiple factors affect the behaviors of demonstrations on the few-shot inference. Intuitively, we propose to view demonstration selection as a combinatorial optimization problem, and introduce exploration-exploitation to iteratively find the optimal policies, which relies on the three-fold framework: 1) Multi-modal Fusion Encoder to catch the rich features interactively; 2) Stochastic Beam Search with auto-regressive decoder to find possible strategies for demonstration combination selection; 3) Responsive Reinforcement to exploit those strategies for finding the optimal policy.

\subsection{Multi-modal Interactive Encoding}
While in vanilla ICL fixed demonstrations can be embedded into prompts for VLMs inference \cite{Brown2020Language}, several works have explored heuristic methods as better strategies, such as selecting according to similarity \cite{wang2022training,zemlyanskiy2022generate} or diversity \cite{Itay2023Diverse,gao2024ambiguityawareincontextlearninglarge}. Although these methods have achieved improvements, they do not cover two major problems that theoretically limit the potential of VLMs: a) These methods do not explore dynamic multimodal relationships, which mainly focus on hand-crafted heuristic strategies. b) Selection of demonstrations is independent, failing to consider the interaction between demonstrations. This is also prone to issues such as information redundancy. 

To tackle the problems, we therefore propose our \textit{Multi-modal Interactive Encoding}, which includes two parts, multi-modal encoder $f_e$ and fusion encoder $f_f$. Given query $(\mathbf t,\mathcal I)$ and candidate demonstrations $\mathcal D$, different modalities are first fed into multi-modal encoder $f_e$ to get the textual feature $\{\mathbf f_{\mathbf t_q},\mathbf f_{\mathbf t_1},\mathbf f_{\mathbf t_2},\dots\}$ and image features $\{\mathbf f_{\mathbf I_q},\mathbf f_{\mathbf I_1},\mathbf f_{\mathbf I_2},\dots\}$, it has the following form:
\begin{align}
 \{\mathbf f_{\mathbf t_q},\mathbf f_{\mathbf t_1},\mathbf f_{\mathbf t_2},\dots\} &= \{f_e(\mathbf t_q),f_e(\mathbf t_1),f_e(\mathbf t_2),\dots\}, \\
 \{\mathbf f_{\mathbf I_q},\mathbf f_{\mathbf I_1},\mathbf f_{\mathbf I_2},\dots\}&=\{f_e(\mathbf I_q),f_e(\mathbf I_1),f_e(\mathbf I_2),\dots\}.
\end{align}

To account for the interaction between multi-modal features, as well as the interaction between candidate demonstrations and queries, we implement $f_f$ as a transformer encoder, thereby ensuring global interactions among tokens. Features of different modalities are first embedded with separate functional embedding $\Delta^f_{\{t,i\}}$ to distinguish them. Subsequently, positional encodings $\Delta^p_i$ are respectively added to features of different modalities, thereby establishing correlations among corresponding features
\begin{align}
 \mathbf f_{{\{\mathbf t,\mathbf I\}}_{q}} &\gets \mathbf f_{{\{\mathbf t,\mathbf I\}}_q}+\Delta^f_{\{t,i\}}, \\
 \mathbf f_{{\{\mathbf t,\mathbf I\}}_k} &\gets \mathbf f_{{\{\mathbf t,\mathbf I\}}_k}+\Delta^f_{\{t,i\}}+\Delta^p_k.
\end{align}
Next, the queries and examples are fed into the fusion encoder $f_f$ to obtain interactive feature $\mathbf M$.
\begin{align}
 \mathbf M=f_f(\begin{bmatrix}
 \mathbf f_{\mathbf t_q} & \mathbf f_{\mathbf I_q} & \mathbf f_{\mathbf t_1} & \mathbf f_{\mathbf I_1} & \mathbf f_{\mathbf t_2} & \mathbf f_{\mathbf I_2} & \cdots
    \end{bmatrix}).
\end{align}
During this process, features of different modalities of queries and candidate demonstrations are fused to learn the characteristics of multi-modal tasks.

\subsection{Stochastic Beam Search Exploration}

\paragraph{Auto-regressive Decoding} Given the black-box nature of both the problem characteristics and VLMs, it is challenging for humans to manually design appropriate demonstration sets and selection strategies. To frame demonstration selection as a combinatorial optimization problem, we introduce an auto-regressive transformer decoder $f_d$.

With $\mathbf M$ by interactive encoder, decoder $f_d$ then predicts the $i$-th demonstration $\mathbf d_i$ based on the queries, candidates and other $i-1$ demonstrations $\mathcal E$ by
\begin{align}
 p_i&=\alpha_t\mathrm{dist}(f_d(\mathbf M,\mathcal E),\mathbf f_{\mathbf t_i})+\alpha_i\mathrm{dist}(f_d(\mathbf M,\mathcal E),\mathbf f_{\mathbf I_i}), \\
 i&={\arg\max}_{i}\left[p_i  
    \leq \min\{p_j\}\right],
\end{align}
where $\mathrm{dist}()$ is the cosine similarity in our implementation, $\alpha_{\{t,i\}}$ are weight coefficients. This process is auto-regressive to achieve the policy $\pi_\theta$
\begin{equation}
    \pi_\theta(\mathcal E \mid \mathbf t_q,\mathcal I_q)=\prod_{i=1}^{m} p(\mathbf d_i \mid \mathbf d_1, \mathbf d_2,\dots,\mathbf d_{i-1},\mathbf t_q,\mathcal I_q).
\end{equation}
Thus the selection is modeled as a combinatorial problem.

\paragraph{Stochastic Beam Search} To facilitate exploration, we incorporate beam search into our algorithm as a method for exploring various policies. Beam Search is a heuristic search algorithm widely adopted in NLP tasks such as sequence generation.
However, due to beam search always expanding the highest probability option at each step, it is not suitable for policy exploration. We modify the traditional beam search to a stochastic sampling version, allowing for stochastic exploration. The algorithm is listed in \Cref{alg:probabilistic-beam-search}. Intuitively, it does not always accept the current ``optimal'' options, thereby introducing the exploration.

Through this algorithm, the model can start from a random policy, continuously explore various strategies, and provide potential optimization directions for the framework.

\begin{algorithm}[t]
    \caption{Stochastic beam search}\label{alg:probabilistic-beam-search}
    \KwData{policy $\pi_\theta$, candidate demonstration $\mathcal D$, size of demonstration combination $m$}
    \KwResult{$B_m$, set of candidate demonstration sets.}
    $B_0 \gets \{(1, \varnothing)\}$\;
    \For{$i \gets 1$ \KwTo $m$}{
        $B \gets \varnothing$\;
        \For{$(s,\mathcal E) \in B_{t-1}$}{
            \Comment{Enumerate and concatenate every demonstration}
            \For{$\mathbf d \in \mathcal D$}{
                $s \gets s \cdot \pi_\theta(\mathbf d \mid \mathcal E,\mathbf t_q,\mathcal I_q)$\;
                $B \gets B \cup \{(s,\mathcal E \cup \mathbf d)\}$\;
            }
        }
        \Comment{Normalize the possibility}
        $S \gets 0$\;
        \For{$(s,\mathcal E) \in B$}{
            $S \gets S + |s|$\;
        }
        $s \gets \frac{s}{|S|}$ for each item in $B$\;
        sample from $B$ with possibility $s$ to get $B_i$\;
    }
\end{algorithm}

\begin{table*}[thbp]
    \caption{Experimental comparison using VQAScore with the state-of-the-art methods on four VQA benchmarks.}
    \label{tab:main-comparison}
    \centering
    \begin{tabular}{c|ccccccccc|c}
    \toprule
    \multirow{2}{*}{\textbf{Methods}} & \multirow{2}{*}{\textbf{OKVQA}} & \multirow{2}{*}{\textbf{TextVQA}} & \multirow{2}{*}{\textbf{Vizwiz}} & \multicolumn{6}{c|}{\textbf{MMStar}} & \multirow{2}{*}{\textbf{Avg.}}\\
    &&&&CP&FP&IR&LR&ST&MA&\\
    \midrule
 zero-shot & $57.5$ & $65.2$ & $32.8$ & $52.0$ & $\mathbf{37.6}$ & $\mathbf{47.6}$ & $25.3$ & $17.9$ & $\mathbf{33.2}$ & $41.0$ \\
 random & $61.1$ & $67.4$ & $40.5$ & $57.2$ & $33.9$ & $46.7$ & $26.6$ & $20.5$ & $28.8$ & $42.5$\\
 similarity (Qwen \cite{bai2023qwenvlversatilevisionlanguagemodel}) & $60.6$ & $67.2$ & $48.7$ & $57.2$ & $31.4$ & $47.2$ & $32.8$ & $26.2$ & $27.1$ & $44.3$ \\
 BM25 \cite{robertson2009probabilistic} & $60.3$ & $66.9$ & $43.6$ & $\mathbf{59.8}$ & $34.1$ & $51.1$ & $30.6$ & $23.6$ & $24.9$ & $43.9$\\
 Ours (Qwen \cite{bai2023qwenvlversatilevisionlanguagemodel}) & $\mathbf{64.7}$ & $\mathbf{67.4}$ & $\mathbf{53.6}$ & $56.8$ & $36.2$ & $46.6$ & $\mathbf{34.1}$ & $\mathbf{24.9}$ & $29.3$ & $\mathbf{45.9}$\\
    \midrule
 similarity (LLaVA \cite{Liu2023llava}) & $22.7$ & $33.6$ & $19.8$ & $48.5$ & $28.4$ & $43.7$ & $33.6$ & $\mathbf{23.6}$ & $27.5$ & $31.3$ \\
 Ours (LLaVA \cite{Liu2023llava}) & $\mathbf{44.3}$ & $\mathbf{50.1}$ & $\mathbf{45.6}$ & $\mathbf{54.6}$ & $\mathbf{38.0}$ & $\mathbf{49.8}$ & $\mathbf{36.7}$ & $20.1$ & $\mathbf{28.4}$ & $\mathbf{40.8}$\\
    \bottomrule
    \end{tabular}
  \end{table*}

\subsection{Responsive Reinforcement Exploitation}

After establishing a sufficiently flexible model and exploration algorithm, we introduce our Responsive Reinforcement Exploitation to continuously optimize the superior policies.

Given a query $(\mathbf t_q,\mathcal I_q)$, the exploration phase first explores and sample multiple candidate demonstration combinations $\mathcal E_1,\mathcal E_2,\dots, \mathcal E_c$ by the stochastic beam search with policy $\pi_\theta$. Based on the combinations, the framework optimizes the policy model $\pi_\theta$ by maximizing the following objective:
\begin{equation}
    \begin{split}
 &\mathbb E_{(\mathbf t_q,\mathcal I_q) \sim \mathcal T,\{\mathcal E_i\}_{i=1}^c \sim \pi_\theta(\mathcal E \mid \mathbf t_q,\mathcal I_q)} \\
  &\qquad=\frac{1}{c}\sum_{i=1}^c \hat A_i \cdot
        \sum_{k=1}^m \log \pi_\theta(\mathbf d_{i,k} \mid \mathbf d_{i,<k},\mathbf t_q,\mathcal I_q).
    \end{split}
\end{equation}
Inspired by policy gradient methods~\cite{kakade2001natural, shao2024deepseekmathpushinglimitsmathematical}, this optimization objective aims to adaptively optimize the joint probability of various demonstration combinations under the policy model. The advantage term $\hat A_i$ represents the relative benefit of each combination, where a larger advantage value indicates a more beneficial combination, thus warranting an increase in its corresponding joint probability.
Specifically, the advantage $\hat A_i$ is responsively calculated as
\begin{align}
 A_i&=r\left(\mathcal F\left(P(\mathbf t_q,\mathcal E_i),\mathcal I_q \cup \bigcup_{\mathcal I_j \in \mathcal E_i} \mathcal I_j\right),\mathbf y\right), \\
 \hat A_i&=\frac{A_i-\mathrm{mean}(\mathbf A)}{\mathrm{var}(\mathbf A)},
\end{align}
where $r(\cdot)$ is reward function and $A_i$ is the original advantages for each combination. $\mathbf A = [A_1,...,A_c]$ denotes the original advantages of all explored combinations for one query. Based on $\mathbf A$, the final advantage $\hat A_i$ is obtained through query-wise standard normalization. It mitigates the distributional discrepancies across queries while ensuring a balanced distribution of positive and negative advantage values.
For our experiments, we simply implement $r(\cdot)$ as
\begin{equation}
 r(\mathbf o,\mathbf y)=\mathds 1_{[\mathbf o~\textrm{is equivalent to}~\mathbf y]},
\end{equation}
where $\mathds 1_{[\textrm{condition}]}$ is the indicator function, equaling $1$ if and only if the output $\mathbf o$ matches the ground-truth $\mathbf y$; otherwise, $0$.
In this way, our approach is able to leverage the explored strategies to gradually uncover adaptive sampling policies that are suitable for various types of queries.

\begin{algorithm}[t]
  \caption{Training Algorithm}\label{alg:training}
  \KwData{task $\mathcal T$, candidate demonstrations $\mathcal D$, size of demonstration sets $m$.}
  \KwResult{policy $\pi_\theta$}
 init $\pi_\theta$ randomly\;

  \For{$(\mathbf t_q,\mathcal I_q,\mathbf y) \sim \mathcal T$}{
    $\{(s_i,\mathcal E_i)\}_{i=1}^c \gets \textrm{beam-search}(\pi_\theta,\mathcal D,m,c)$\;
    $\mathbf o_i \gets \mathcal F(P(\mathbf t_q,\mathcal E_i),\bigcup_{\mathcal I_j \in \mathcal E_i} \mathcal I_j)$\;
    $A_i \gets r(\mathbf o_i,\mathbf y)$\;
    $\hat A_i \gets \frac{A_i-\mathrm{mean}(\mathbf A)}{\mathrm{var}(\mathbf A)}$\;
 Optimize $\pi_\theta$ to maximize $\mathcal L$\;
 }
\end{algorithm}

\section{Experiments}

\subsection{Experiment Setups}

\noindent{}\textbf{Datasets.}
To demonstrate the generality of our method, following the previous studies \cite{bai2023qwenvlversatilevisionlanguagemodel}, we conduct experiments on four representative VQA benchmarks, \ie, OKVQA \cite{Marino2019okvqa}, TextVQA \cite{singh2019towards}, Vizwiz \cite{Bigham2010VizWiz}, and MMStar \cite{chen2024we}. 
OKVQA is a large-scale VQA benchmark constructed based on the MS-COCO dataset, which contains 14k question-answer pairs and 14k images. TextVQA dataset contains 28k images and associated questions with human-annotated answers. Vizwiz dataset contains 24k image/question pairs collected from the blind. MMStar is designed to evaluate the multi-modal capacities with carefully balanced challenge samples.
These datasets include images and diverse text questions, reflecting ICL performance and showcasing the adaptability of methods to various tasks.

\noindent{}\textbf{Baseline and Evaluation Metrics.}
We adopt a heuristic similarity policy as baseline, which selects demonstrations by cosine similarities of the embeddings. For fair comparisons, we follow previous studies \cite{bai2023qwenvlversatilevisionlanguagemodel} to use VQAScore as the main metric. The results are reported on validation sets.

\subsection{Implementation Details}

\noindent{}\textbf{Data Processing.}
Following previous studies \cite{bai2023qwenvlversatilevisionlanguagemodel}, we adopt the official validation sets of OKVQA, TextVQA, and Vizwiz benchmarks to evaluate the performance. As no splits are provided for MMStar, following convention, we divide it into training and validation sets in a 7:3 ratio, treating the training sets as candidate demonstrations for both training and validation sets.

\noindent{}\textbf{Training Details.}
We adopt ImageBind \cite{girdhar2023imagebind} as $f_e$ in the interactive encoding. We use Qwen-VL \cite{bai2023qwenvlversatilevisionlanguagemodel} and LLaVA \cite{Liu2023llava} as LVLMs. We implement $\Delta$ as learnable functional and positional encoding following \cite{carion2020end}. AdamW optimizer is adopted to train models for $10$ epochs. The learning rate is $5 \times 10^{-5}$. The batch size is set as $8$. The hyperparameter $c$ in responsive reinforcement is set as $4$. The beam width and temperature are set as $9$ and $1.0$. The selection model consists of $6$-layer transformers with $8$ attention heads and $2048$ hidden dimensions. As we evaluate LVLMs locally, all experiments are conducted on 8 NVIDIA A800 GPUs.

\begin{table}[t]
  \caption{Ablation studies of different modules on the OKVQA benchmark. \textbf{SBS}: Stochastic beam search. \textbf{AR}: Auto-regressive. \textbf{RR}: Responsive Reinforcement.}
  \label{tab:component-ablation}
  \centering
  \begin{tabular}{ccc|cc}
  \toprule
  \textbf{Training Alg.} & \textbf{SBS} & \textbf{AR} & \textbf{VQAScore} & $\Delta$ \\
  \midrule
 RR & $\checkmark$ & & $63.3$ & {\color{red}$1.4\downarrow$} \\
 RR & & $\checkmark$ & $63.3$ & {\color{red}$1.4\downarrow$} \\
 Preference & $\checkmark$ &  $\checkmark$ & $62.9$ & {\color{red}$1.8\downarrow$} \\
  \midrule
 RR & $\checkmark$ & $\checkmark$ & $64.7$ & $-$ \\
  \bottomrule
  \end{tabular}
\end{table}

\subsection{Comparison with Representative Methods}

We respectively compare our methods with four methods, zero-shot, random, similarity, BM25, on four datasets including OKVQA \cite{Marino2019okvqa}, TextVQA \cite{singh2019towards}, Vizwiz \cite{Bigham2010VizWiz}, and MMStar \cite{chen2024we}. \Cref{tab:main-comparison} show the comparisons of VQAScore between ours and other methods. Our method surpasses other representative methods on four datasets. Especially on OKVQA and VizWiz, it improves significantly, increasing by $4.1$ and $4.9$ compared to \textit{similarity}. The average performance is improved by $1.6$. Additionally, it is noteworthy that our method achieves improvements in fine-grained perception compared to coarse perception, as well as in logical reasoning compared to instance reasoning. For coarse perception and instance reasoning tasks, directly selecting similar demonstrations yields better performance, whereas our proposed reinforcement method is unstable in fitting this fixed strategy.
Furthermore, \Cref{tab:main-comparison} shows our method's lower performance
on math tasks. The math problems follow rigid rules where minor variations can lead to completely different solution paths. Related but non-isomorphic demonstrations interfere with LVLMs' understanding.
In general, the experimental results show that our method is more effective than various heuristic approaches. It provides more significant improvement for weaker models \cite{chen2024we}, \eg, LLaVA. By treating the selection as a combinatorial optimization problem, we achieve better results than the independent selection strategy.

\subsection{Performance Analysis}

\textbf{Ablation Studies of Components.} To verify the effectiveness, we conduct extensive ablation experiments. The results are shown in \Cref{tab:component-ablation}. Without auto-regressive generation, the selection of each demonstration is independent, leading to a significant decrease in performance. This indicates that the interactions between demonstrations play a crucial role in analogical understanding. Stochastic beam search not employed, the limited self-exploration restricts the variety of strategies and declines the performance. When using preference loss, the optimization is not stable to converge to the better solutions found by responsive reinforcement. Our proposed framework shows strong performance in promoting in-context learning.

\begin{figure}[t]
  \centering
  \includegraphics[width=1\linewidth]{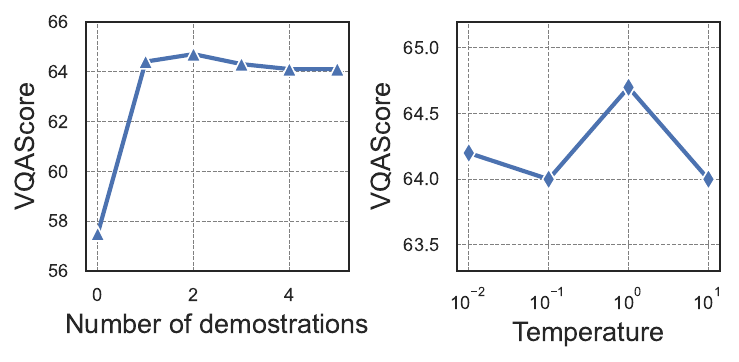}
  \caption{\textbf{Left}: VQAScore on OKVQA benchmark of our method with different number of demonstrations. \textbf{Right}: VQAScore on OKVQA benchmark of our method with different temperature values in stochastic beam search.}
    \label{fig:ablation-demo-size}
\end{figure}

\noindent{}\textbf{Effects of Numbers of Demonstrations.} The number of demos impacts ICL. We conduct experiments to investigate the contribution of demo sizes to the performance. From \Cref{fig:ablation-demo-size}(left), it can be observed that as the number increases, the performance gradually improves. However, when the demonstration set becomes too large, performance shows a certain decline. We believe that while a larger number of demonstrations provides more information, it challenges the reinforcement learning process, thereby decreasing the performance. Additionally, too many demonstrations also add a number of tokens, raising inference costs. Therefore, in our method, we set the size to $2$.

In beam search, temperature controls the gap between high- and low-probability demonstrations during sampling, thus affecting policy exploration. We conduct experiments with several different temperatures. The performance in \Cref{fig:ablation-demo-size}(right) shows a general trend. It first rises then declines as temperature increases, thus we set temperature as $1$. This hyper-parameter should be adjusted based on the characteristics of the downstream tasks.

\begin{figure}[t]
  \centering
   \includegraphics[width=.9\linewidth]{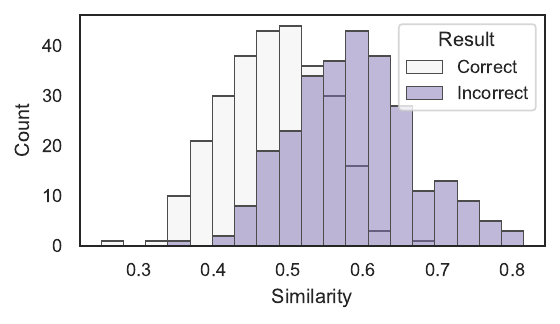}
   \caption{The similarity distribution of demonstrations that correct wrong predictions on the OKVQA benchmark. The demonstrations selected with ours are dissimilar to the queries but correct the predictions.}
   \label{fig:strategies-vs-similarity}
\end{figure}

\begin{figure}[t]
  \centering
   \includegraphics[width=.8\linewidth]{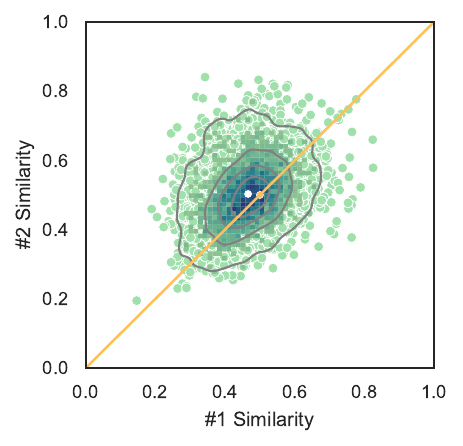}
   \caption{The relationships in cosine similarity of two demonstrations selected with ours on the OKVQA dataset. The yellow dot and white dot represent the centroid $(0.5,0.5)$ and the data centroid, respectively. The data exhibits an overall leftward shift.
   }
   \label{fig:similarity-vs-similarity}
\end{figure}

\begin{figure*}[t]
  \centering
   \includegraphics[width=.9\linewidth]{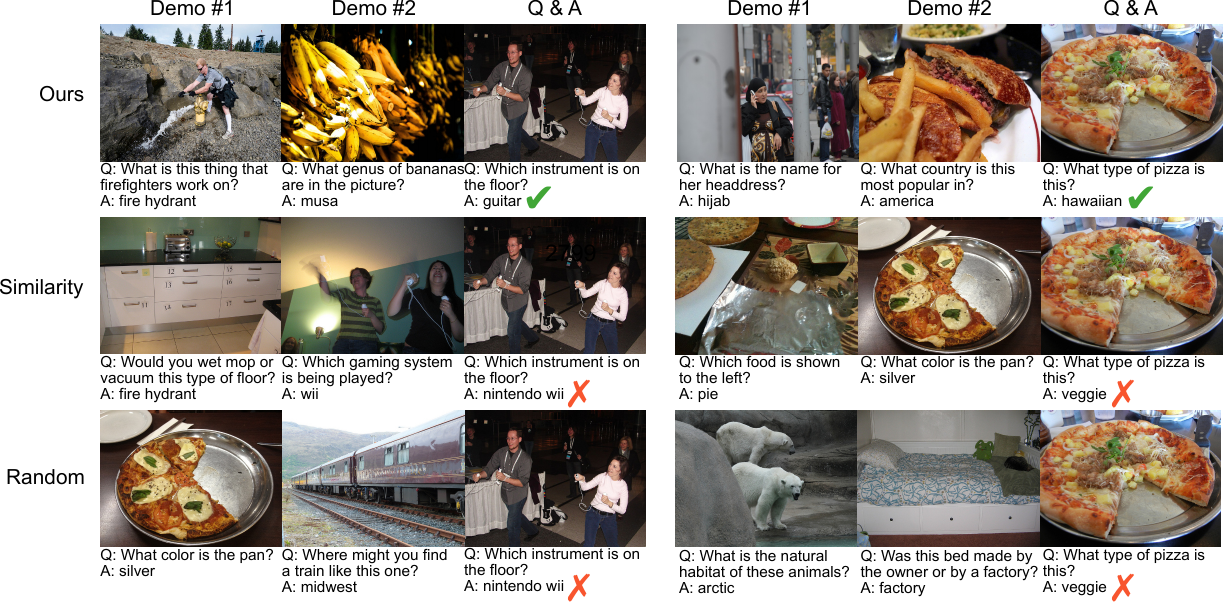}
   \caption{Demonstrations selected with our method, similarity, and random strategies on the OKVQA dataset.
   The demos selected based on criteria such as similarity are closer to queries, but differ in structure or in key objects, which are ineffective in analogical reasoning.
   Our approach adaptively learns to identify and utilize demonstration combinations that maximize both complementarity and utility. 
   }
   \label{fig:visualization}
\end{figure*}

\noindent{}\textbf{Inefficient Similarity Strategy.} For the similarity-based strategy, we conduct more detailed experiments to verify its effects. In \Cref{fig:strategies-vs-similarity}, we illustrate the queries corrected by our method on the OKVQA dataset. Our method selects demonstrations with lower similarity for these examples and predicts the correct answer; the similarity-based strategy, on the other hand, selects more similar demonstrations but misleads the predictions. In this manner, we verify that similar demonstrations are not necessarily the most effective.
We observe the similarity relationships between two selected demonstrations. As shown in \Cref{fig:similarity-vs-similarity}, our method tends to select a dissimilar demonstration first, followed by a similar one. This suggests a complex selection strategy that maintains both similarity and diversity.
Next, we observe the demonstration ordering as many recent works \cite{lu2022fantastically, wu2023self, guo2024makes} treat it as one main factor affecting efficiency. The results in \Cref{tab:permutation} suggest that ordering effects are specific to certain strategies. Only similarity-based strategy is affected and improved when the query is closer to a more similar demo. The results indicate that our method captures robust characteristics of similarity and diversity.

\begin{table}[t]
    \centering
    \caption{The effect of the orders of demonstrations on the OKVQA benchmark.}
    \label{tab:permutation}%
    \begin{tabular}{c|cc|c}
    \toprule
    \textbf{Methods}&\textbf{normal perm.} & \textbf{reversed perm.} & $\Delta$\\
    \midrule
 Similarity & $60.63$ & $60.92$ & $+0.29$ \\
 Ours & $64.66$ & $64.69$ & $+0.03$ \\
    \bottomrule
    \end{tabular}%
\end{table}%

\noindent{}\textbf{What Makes A Better Demonstrations?}
\Cref{tab:main-comparison} indicates that policies based on human intuition, such as \textit{rank by similarity}, are not optimal.
Then what makes a better demonstration? To verify this issue, we conduct an in-depth examination of the policies learned by models. The results are shown in \Cref{fig:visualization}.
In the first set of images, the similarity strategy correctly selects indoor scenes and gameplay images. However, the query does not address the main object of the image. The similar demonstrations fail to provide relevant information but mislead LVLMs, causing them to focus on irrelevant salient objects. The random strategy further demonstrates that LVLMs are easily disturbed by salient objects. In contrast, the demonstrations selected by our method direct the attention to small background instruments.
Similarly, in the second set of images, the similarity strategy selects two food images according to the query. However, it does not lead to effective inference for query questioning the types. Random strategy reveals that LVLMs tend to predict incorrect answers. The demonstrations selected by ours prompt LVLMs to focus on the type characteristics.
The use of a single or heuristic strategy fails to account for the multiple factors, including queries, semantics, and image content. Our proposed reinforcement learning framework allows for a more comprehensive exploration of effective policies.

\section{Conclusions and Limitations}

\textbf{Conclusion.} In this paper, we focus on the problem of demonstration selection in in-context learning. We view the selection from a new perspective considering it as a combinatorial optimization problem to model the interactions among the demonstrations. In our framework, we propose multi-modal interactive encoding to catch the relationships between different modalities of queries and candidate demonstrations. We propose auto-regressive decoder with stochastic beam search to enable exploring policies for selection, and responsive reinforcement to exploit them.  Experimental results demonstrate our proposed framework achieves superior performance.

\noindent \textbf{Limitations.} Our proposed method explores language-vision multimodal ICL only. As research progresses, LLMs are incorporating more modalities such as video and audio, and the exploration of ICL within them remains a pressing area for investigation, leaving for our future work.

{
    \small
    \bibliographystyle{ieeenat_fullname}
    \bibliography{main}
}


\end{document}